\documentclass{article} 
\usepackage{iclr2027_conference,times}

\usepackage{amsmath,amsfonts,bm}

\def\eqref#1{equation~\ref{#1}}

\def\1{\bm{1}}

\DeclareMathAlphabet{\mathsfit}{\encodingdefault}{\sfdefault}{m}{sl}
\SetMathAlphabet{\mathsfit}{bold}{\encodingdefault}{\sfdefault}{bx}{n}

\usepackage{graphicx}
\usepackage{listings}
\usepackage{hyperref}
\usepackage{url}
\usepackage[capitalize,noabbrev]{cleveref}

\lstdefinestyle{diffpython}{
    language=Python,
    basicstyle=\ttfamily\footnotesize,
    frame=single,
    columns=fullflexible,
    keepspaces=true,
    showstringspaces=false
}

\lstdefinestyle{bpco}{
    language=Python,
    basicstyle=\ttfamily\footnotesize,
    frame=single,
    columns=fullflexible,
    keepspaces=true,
    showstringspaces=false,
    numbers=left,
    numberstyle=\scriptsize\color{gray},
    xleftmargin=1.5em,
    framexleftmargin=1.2em,
    aboveskip=0.6em,
    belowskip=0.2em
}

\title{Best Practice Critic Optimization}

\author{
  \textbf{Penghui Qi}$^{1}$,
  \textbf{  Xiangxin Zhou}$^{2}$,
  \textbf{ Wee Sun Lee}$^{1}$ \\
  $^1$National University of Singapore \quad $^2$Tencent Hunyuan \\
  \texttt{\{penghuiq,leews\}@comp.nus.edu.sg}
}

\iclrfinalcopy 
\begin{document}

\maketitle

\begin{abstract}
Group-based reinforcement learning methods such as GRPO for large language models avoid training a critic by sampling multiple responses for each prompt. A reliable critic could instead estimate token-level advantages from one response, but standard critic-based training recipes are often unstable. We study this instability and develop \textbf{Best Practice Critic Optimization (BPCO)}\footnote{The BPCO name also reflects its \textbf{B}ounded \textbf{P}rivileged \textbf{C}ritic \textbf{O}ptimization design.}, a recipe that combines DPPO, value predictions bounded to the reward range, Monte Carlo value targets, unnormalized policy advantages, and length-adaptive generalized advantage estimation. Because the critic is used only during training, BPCO can also condition it on reward-defining information, such as a reference answer or grading rubric, that is hidden from the policy. Controlled experiments isolate the effect of each design choice. Across mathematical reasoning tasks with models ranging from 1.5B parameters to 30B-A3B mixtures of experts, BPCO improves a strong critic-based baseline consistently, and matches or exceeds a group-based baseline while sampling one response per prompt. The same recipe also improves learning with rubric-based rewards. These results show that a carefully designed critic provides a reliable alternative to group-relative advantage estimation. Code is available at \url{https://github.com/QPHutu/golden_critic}.
\end{abstract}

\section{Introduction}

Reinforcement learning (RL) has become a standard approach for improving the reasoning and instruction-following abilities of large language models (LLMs)~\citep{ouyang2022training,deepseek_r1,brpo}. Effective RL depends on assigning credit to the sampled tokens~\citep{sutton2018rlbook}. Group-based methods such as GRPO estimate this signal by sampling several responses for each prompt and comparing their rewards~\citep{shao2024deepseekmath,dr.grpo}. This approach avoids training a value function, but it uses multiple rollouts per prompt and assigns the same outcome-based advantage to every token in a response.

A learned critic offers a direct alternative~\citep{schulman2017ppo}. By estimating the expected return of each response prefix, a critic can construct token-level advantages from one rollout~\citep{schulman2016gae,hou2026sao}. In practice, however, critic-based LLM training remains fragile. PPO's ratio clipping treats low- and high-probability tokens unevenly~\citep{qi2026dppo}. Bootstrapped value targets can inherit critic error~\citep{yuan2025vcppo}, and a fixed GAE parameter gives the terminal reward very different weights in short and long responses~\citep{yue2025vapo}. We identify two additional mismatches in common implementations. First, a linear value head can predict outside the known range of the return. Second, batch-wise advantage normalization forces every batch to have unit-scale advantages, even when the residual policy signal has become small. Our controlled study shows that both choices can destabilize training.

A critic also creates an opportunity that group-relative estimators do not directly exploit. Because the critic is discarded after training, it may receive reward-defining information that is unavailable to the policy. Examples include a reference answer or official solution in mathematical reasoning and a prompt-specific rubric in rubric-based evaluation. Such information is determined by the prompt and therefore does not change the ideal value function. Presenting it explicitly can nevertheless make that function easier to approximate, without changing the policy's inputs or deployment requirements.

We combine these choices into Best Practice Critic Optimization (BPCO), a single-rollout actor--critic recipe. BPCO uses DPPO~\citep{qi2026dppo} to define clipping in terms of the sampled token's probability change. It bounds value predictions to the reward range and trains the critic directly on observed outcome rewards. For the policy update, it preserves the scale of the raw advantages and adapts the GAE parameter to response length. BPCO can additionally use reward-defining information as privileged critic input when such information is available. Together, these choices align the critic's output, target, and inputs with the policy signal it produces.

We develop the recipe incrementally in a controlled sanity test (\Cref{sec:sanity_test}), where failure to fit a small solvable dataset reveals optimization problems. We then evaluate BPCO on a 40.3K-problem mathematical dataset (\Cref{sec:larger_dataset}), two 30B-A3B mixture-of-experts models (\Cref{sec:larger_model}), and a rubric-reward task (\Cref{sec:rubrics}). The experiments support three findings. First, BPCO improves the critic-based baseline across model and dataset scales. Second, privileged information can accelerate critic learning, but its policy benefit depends on the task and the degree of overfitting. Third, BPCO matches or exceeds a group-based baseline while using one response per prompt. These results establish a practical recipe for single-rollout critic-based LLM RL.

\section{Background}
\label{sec:background}

\subsection{Proximal Policy Optimization}
\label{sec:ppo_loss}

Given a prompt $x$, a language model with parameters $\theta$ generates a response $y=(y_1,\ldots,y_T)$ autoregressively. At step $t$, the state is the prefix $s_t=(x,y_{<t})$, the action is the next token $y_t$, and the policy is $\pi_\theta(y_t\mid s_t)$. We consider outcome rewards: a completed response receives a scalar reward $R(x,y)$, and all intermediate rewards are zero.

Proximal Policy Optimization (PPO) uses a clipped surrogate objective~\citep{schulman2017ppo}. Let $\mu$ be the behavior policy that generated the rollouts, and define sampled-token probability ratio as
\begin{equation*}
    \rho_t(\theta)=\frac{\pi_\theta(y_t\mid s_t)}{\mu(y_t\mid s_t)}.
\end{equation*}
Given an advantage estimate $\widehat{A}_t$, PPO maximizes
\begin{equation}
    \mathcal{L}_{\mathrm{PPO}}(\theta)=
    \mathbb{E}_t\left[\min\left(\rho_t(\theta)\widehat{A}_t,
    \operatorname{clip}(\rho_t(\theta),1-\epsilon,1+\epsilon)\widehat{A}_t\right)\right].
    \label{eq:ppo-objective}
\end{equation}
The clipped term removes the incentive to move the ratio farther beyond the clipping boundary in the direction favored by $\widehat{A}_t$, forming a trust region to stabilize training~\citep{schulman2015trust}.

\subsection{Divergence Proximal Policy Optimization}
\label{sec:dppo_loss}

PPO applies the same ratio threshold to every token. In a large vocabulary, this rule clips small absolute changes to low-probability tokens while allowing much larger absolute changes to high-probability tokens~\citep{qi2026dppo}. Divergence Proximal Policy Optimization (DPPO) instead defines the clipping boundary in terms of the sampled token's probability change. The binary total-variation variant used in this work replaces $\epsilon$ in \Cref{eq:ppo-objective} with $\epsilon/\mu(y_t\mid s_t)$:
\begin{equation}
    \mathcal{L}_{\mathrm{DPPO}}(\theta)=
    \mathbb{E}_t\left[\min\left(\rho_t(\theta)\widehat{A}_t,
    \operatorname{clip}\left(\rho_t(\theta),1-\frac{\epsilon}{\mu(y_t\mid s_t)},1+\frac{\epsilon}{\mu(y_t\mid s_t)}\right)\widehat{A}_t\right)\right].
    \label{eq:dppo-objective}
\end{equation}
Equivalently, DPPO constrains the probability shift of the sampled token under the policy update, i.e.,
$\left|\pi_\theta(y_t \mid s_t)-\mu(y_t \mid s_t)\right| \leq \epsilon$. This gives sampled tokens a common absolute-probability threshold rather than a common ratio threshold.

\subsection{Critic-Based Methods}
\label{sec:critic}

Critic-based methods estimate the expected return of each prefix. For rollouts from $\mu$, the value function is
\begin{equation*}
    V^\mu(s_t)=\mathbb{E}_{\mu}\!\left[R(x,y)\mid s_t\right],
\end{equation*}
and the critic $V_\phi(s_t)$ approximates this quantity. Let $\phi_{\mathrm{old}}$ denote the frozen critic parameters used to construct targets. Generalized advantage estimation (GAE)~\citep{schulman2016gae} first computes temporal-difference residuals and then forms an exponentially weighted sum:
\begin{align}
    \delta_t &= r_t+\gamma V_{\phi_{\mathrm{old}}}(s_{t+1})-V_{\phi_{\mathrm{old}}}(s_t),
    \label{eq:td-error}\\
    \widehat{A}^{\mathrm{GAE}(\lambda)}_t
    &=\sum_{l=0}^{T-t}(\gamma\lambda)^l\delta_{t+l}.
    \label{eq:gae}
\end{align}
Here $r_t=0$ for $t<T$, $r_T=R(x,y)$, and $V_{\phi_{\mathrm{old}}}(s_{T+1})=0$. In LLM training, $\gamma=1$ is commonly used. The value of $\lambda$ controls the degree of bootstrapping. Smaller $\lambda$ can reduce variance but makes the estimate biased and more sensitive to critic error. With $\lambda=1$, the sum telescopes to $R(x,y)-V_{\phi_{\mathrm{old}}}(s_t)$, which is an unbiased Monte Carlo estimate without bootstrapping.

Many implementations construct the critic target as
\begin{equation}
    \widehat{V}_t(\lambda)=\widehat{A}^{\mathrm{GAE}(\lambda)}_t+V_{\phi_{\mathrm{old}}}(s_t)
    \label{eq:value-target}
\end{equation}
and minimize
\begin{equation}
    \mathcal{L}_{\mathrm{V}}(\phi)=\mathbb{E}_t\left[\left(V_\phi(s_t)-\widehat{V}_t(\lambda)\right)^2\right].
    \label{eq:critic-loss}
\end{equation}
The policy update uses $\widehat{A}^{\mathrm{GAE}(\lambda)}_t$ in \Cref{eq:ppo-objective} or \Cref{eq:dppo-objective}.

\subsection{Group-Based Methods}

Group-based methods avoid a critic by sampling $G$ responses $\{y^{(i)}\}_{i=1}^{G}$ for each prompt~\citep{shao2024deepseekmath}. Let $R_i=R(x,y^{(i)})$, and let $\mu_R$ and $\sigma_R$ be the mean and standard deviation of the $G$ rewards. GRPO assigns every token in response $i$ the advantage
\begin{equation}
    \widehat{A}^{\mathrm{GRPO}}_{t,i}=\frac{R_i-\mu_R}{\sigma_R}.
    \label{eq:group-advantage}
\end{equation}
Dr.~GRPO removes the standard-deviation normalization, which can otherwise reweight prompts according to their within-group reward variance~\citep{dr.grpo}. Its advantage is
\begin{equation}
    \widehat{A}^{\mathrm{Dr.GRPO}}_{t,i}=R_i-\mu_R.
    \label{eq:drgrpo-advantage}
\end{equation}

\subsection{Critics for Long-Response LLM Reinforcement Learning}

Recent work has revisited critics for long-response LLM RL\@. VC-PPO decouples the GAE parameters used for policy and critic training and pretrains the critic to reduce initialization error~\citep{yuan2025vcppo}. VAPO combines these ideas with length-adaptive GAE and other long-response optimization techniques~\citep{yue2025vapo}. SAO uses a critic for single-rollout asynchronous RL, with additional critic updates and frozen attention to track a changing policy~\citep{hou2026sao}. BPCO is complementary: it isolates the effects of the critic's output range, batch-wise advantage normalization, and optional reward-defining inputs in a single-rollout setting.

\section{Building BPCO: A Controlled Study}
\label{sec:sanity_test}

We begin from a \href{https://github.com/verl-project/verl/tree/86e8123643c982343b11fa21dc1df96ab07ced3d}{verl commit from June 16, 2026} and study critic stability in a controlled sanity test~\citep{qi2025defeating,qi2026dppo}. We fine-tune DeepSeek-R1-Distill-Qwen-1.5B~\citep{deepseek_r1} on 1,460 mathematical problems that the initial model can solve. A suitable training recipe should fit this deliberately small dataset to nearly 100\% reward. Failure to do so exposes an optimization problem rather than a lack of model capacity or reward signal.

Each iteration contains 1,024 trajectories. We use a minibatch size of 256 and one optimization epoch, giving four optimizer minibatches per iteration. Following the verl defaults~\citep{verl}, the policy and critic learning rates are $10^{-6}$ and $10^{-5}$, respectively. We observed no benefit from critic warm-up in this small-data setting and therefore update the policy and critic from the first iteration. Each run lasts 1,500 iterations. Because fitting this small dataset can harm generalization, we monitor AIME 2025 avg@32, the mean accuracy over 32 sampled responses per problem, as a held-out metric.

We modify the default recipe one component at a time. The starting point uses PPO (\Cref{sec:ppo_loss}), standard GAE and critic targets with $\lambda=1$ (\Cref{sec:critic}). Unless stated otherwise, each step retains all preceding changes.

\begin{figure}[h]
    \centering
    \includegraphics[width=\textwidth]{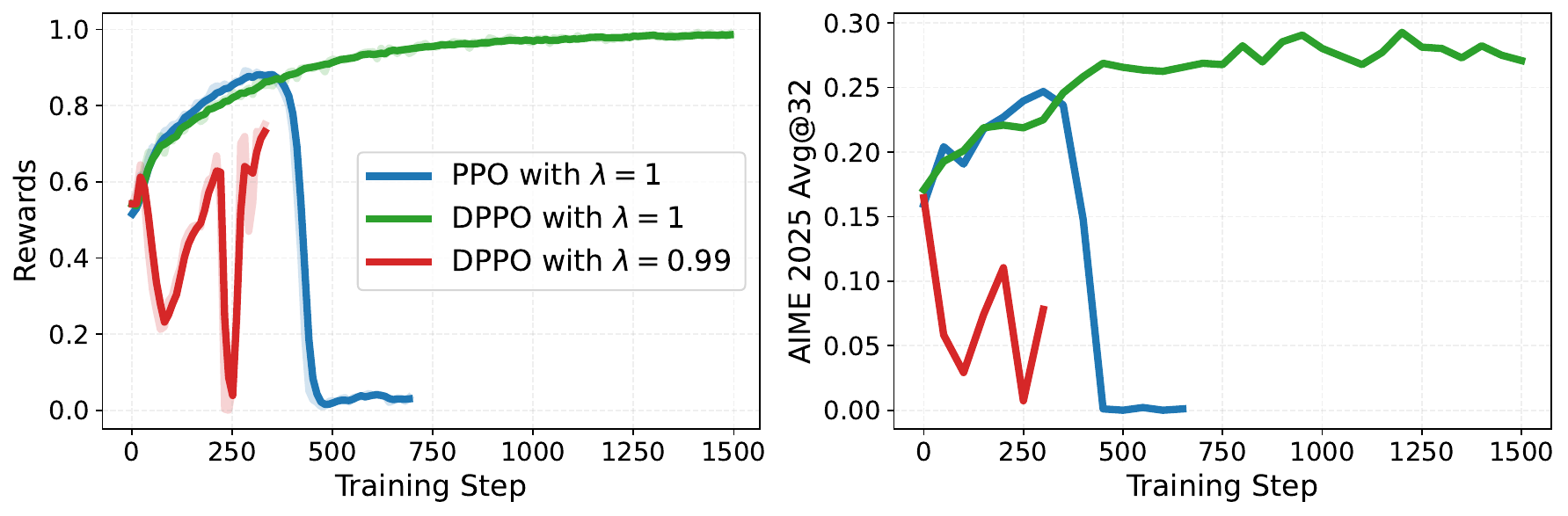}
    \caption{PPO and DPPO in the sanity test. With $\lambda=1$, PPO's training reward collapses after an initial increase, whereas DPPO remains stable. Using $\lambda=0.99$ makes DPPO unstable again.}
    \label{fig:ppo-vs-dppo-lambda}
\end{figure}

\subsection{Step 1: Replacing PPO with DPPO}
\label{sec:ppo2dppo}

With the PPO objective in \Cref{eq:ppo-objective}, the training reward collapses. Replacing it with the DPPO objective in \Cref{eq:dppo-objective} yields stable optimization when $\lambda=1$, as shown in \Cref{fig:ppo-vs-dppo-lambda}.

Reducing the GAE parameter to $\lambda=0.99$ makes DPPO unstable again. When $\lambda<1$, the policy advantage contains bootstrapped critic predictions. Unless $V_\phi(s_t)=V^\mu(s_t)$ for every visited state, approximation error biases the advantage estimate relative to the Monte Carlo estimator obtained with $\lambda=1$. We therefore use $\lambda=0.99$ as a stress test in the next steps: stabilizing this setting requires the recipe to control how critic error enters the policy update.

\subsection{Step 2: Bounding Values to the Reward Range}

\begin{figure}[h]
    \centering
    \includegraphics[width=\textwidth]{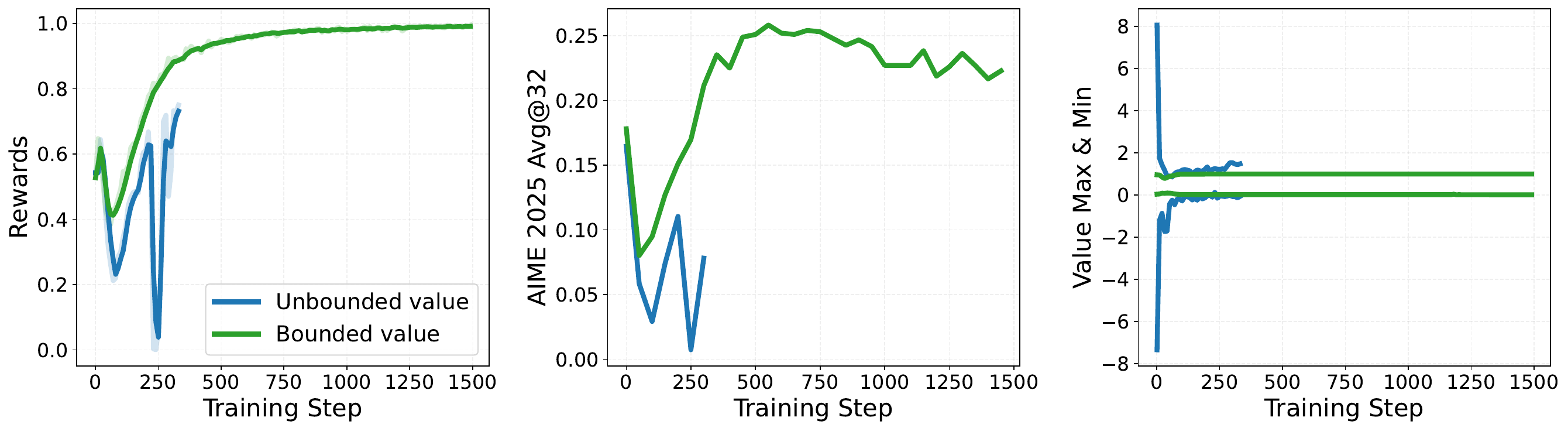}
    \caption{Effect of bounding critic predictions to the reward range. The unbounded linear head predicts values outside the binary-reward range $[0,1]$ (right), leading to unstable training reward (left) and AIME 2025 avg@32 (middle). The bounded value keeps predictions within the reward range and yields stable training.}
    \label{fig:bounded-value}
\end{figure}

Most existing recipes use a linear head to predict the value directly, which is unbounded even when the return is known to lie in a finite interval. Let $[R_{\min},R_{\max}]$ be the reward range and let $z_\phi(s_t)\in\mathbb{R}$ be the linear head output. Because the expectation of a bounded random variable lies in the same interval, a valid value prediction must satisfy $V_\phi(s_t)\in[R_{\min},R_{\max}]$.
We enforce this property with a scaled arctangent:
\begin{equation}
    V_\phi(s_t)=R_{\min}+(R_{\max}-R_{\min})
    \left(\frac{1}{2}+\frac{1}{\pi}\arctan\bigl(z_\phi(s_t)\bigr)\right).
    \label{eq:bounded-value}
\end{equation}
This parameterization maps every finite head output to the open interval $(R_{\min},R_{\max})$ and approaches either endpoint asymptotically. The sanity test uses binary rewards, so $R_{\min}=0$ and $R_{\max}=1$. Empirically, it removes the extreme values produced by the linear head and allows the training reward to approach one (\Cref{fig:bounded-value}).

\subsection{Step 3: Using Unbiased Monte Carlo Value Target}
\label{sec:decoupled_gae}

\begin{figure}[h]
    \centering
    \includegraphics[width=\textwidth]{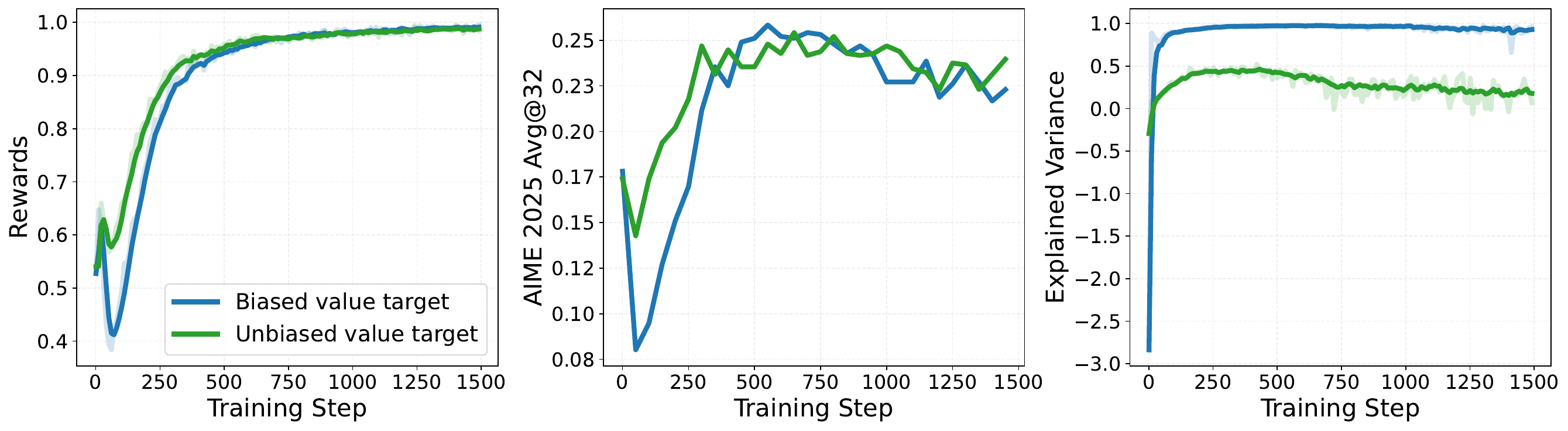}
    \caption{Effect of the unbiased value target. Regressing to the final reward gives more stable training reward (left) and AIME 2025 avg@32 (middle) than a target bootstrapped with $\lambda=0.99$. Explained variance against the bootstrapped target rapidly approaches one (right) because that target is biased; this unexpectedly high value does not imply accurate prediction of the observed return.}
    \label{fig:unbiased-value-target}
\end{figure}

We track how much target variance the critic explains using
\begin{equation}
    \operatorname{EV}(V_\phi,\widehat{V})
    =1-\frac{\operatorname{Var}\!\left(\widehat{V}_t-V_\phi(s_t)\right)}
    {\operatorname{Var}\!\left(\widehat{V}_t\right)}.
    \label{eq:explained-variance}
\end{equation}
An explained variance near one normally indicates a close fit to the chosen target. For the standard target in \Cref{eq:value-target}, however, $\widehat{V}_t(\lambda)$ contains $V_{\phi_{\mathrm{old}}}$ whenever $\lambda<1$. This self-referential target can be easy for the updated critic to predict even when it is inaccurate with respect to the observed return. In \Cref{fig:unbiased-value-target}, explained variance against the bootstrapped target rapidly approaches one while policy training remains unstable.

Following decoupled GAE in VC-PPO~\citep{yuan2025vcppo}, we use separate parameters for the policy advantage and the critic target. We retain $\lambda_{\pi}=0.99$ for the policy, but set $\lambda_V=1$ for critic training. With $\gamma=1$ and outcome-only rewards, the target telescopes to
\begin{equation}
    \widehat{V}_t
    =\widehat{A}^{\mathrm{GAE}(1)}_t+V_{\phi_{\mathrm{old}}}(s_t)
    =R(x,y).
    \label{eq:unbiased-value-target}
\end{equation}
For a continuation sampled from $\mu$, the final outcome is an unbiased Monte Carlo sample of $V^\mu(s_t)$. Decoupling the estimators retains the variance reduction of $\lambda_{\pi}<1$ for the policy while removing bootstrapping bias for the critic target. This change improves reward stability and convergence speed in \Cref{fig:unbiased-value-target}; the reported explained variance is now measured against the observed outcome reward, and becomes reasonable.

\subsection{Step 4: Removing Advantage Normalization}
\label{sec:adv_norm}

\begin{figure}[h]
    \centering
    \includegraphics[width=\textwidth]{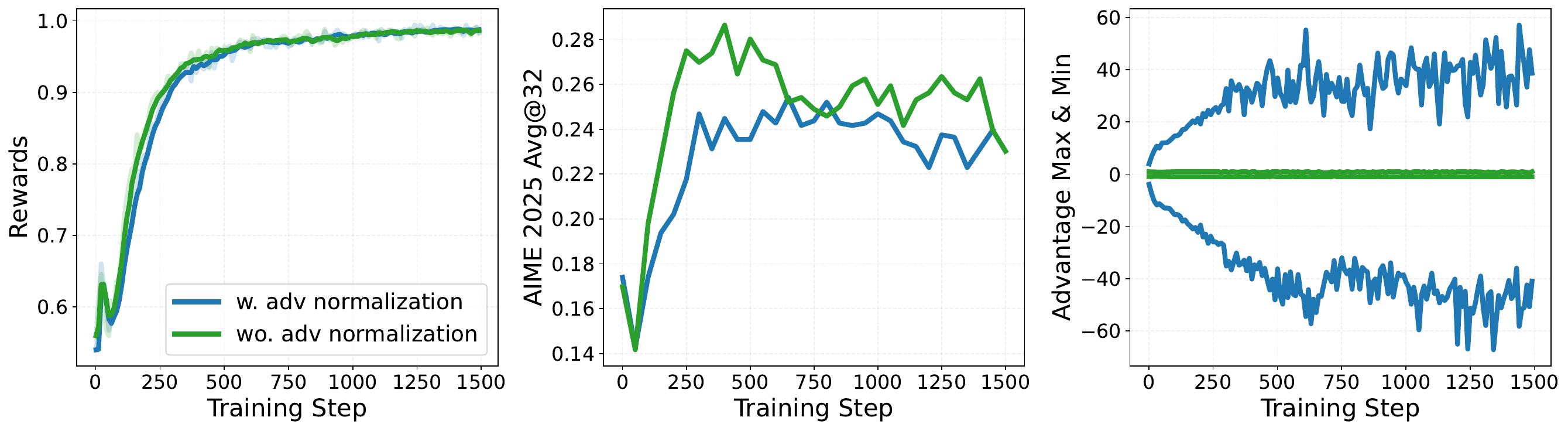}
    \caption{Effect of removing batch-wise advantage normalization. Removing normalization achieves comparable training reward (left) while mitigating the overfitting risk on AIME 2025 (middle). Batch-wise advantage normalization forces small residual advantages to a large, growing range when the policy approaches to optimal (right).}
    \label{fig:advantage-normalization}
\end{figure}

Many PPO implementations normalize advantages within each batch before the policy update \citep{shengyi2022the37implementation}. Let $\bar{A}$ and $\sigma_A$ denote the batch mean and standard deviation of the estimated advantages, respectively. This procedure replaces $\widehat{A}_t$ in \Cref{eq:ppo-objective} and \Cref{eq:dppo-objective} with
\begin{equation}
    \widetilde{A}_t = \frac{\widehat{A}_t-\bar{A}}{\sigma_A}.
    \label{eq:advantage-normalization}
\end{equation}
We find this transformation fundamentally problematic. As the policy approaches optimality, the advantages and their variance should both decrease toward zero, causing the policy update to diminish naturally and preserve the policy. Dividing by $\sigma_A$, however, removes this desirable behavior. When the standard deviation is small, estimation noise is amplified into a large training signal, preventing the update from vanishing near the optimum. Moreover, subtracting $\bar{A}$ can reverse the sign of examples with positive advantages that are smaller than the batch mean, thereby hampering exploration.

We therefore remove batch advantage normalization and use the raw GAE estimates for policy updates. As shown in \Cref{fig:advantage-normalization}, removing normalization keeps the advantage range small and stable, whereas normalization causes the magnitude of the normalized advantages to increase during training. Notably, removing advantage normalization also improves validation performance, which is consistent with the less aggressive updates obtained after the training set has nearly been fit.

\subsection{Step 5: Providing Privileged Information to the Critic}
\label{sec:privileged}

\begin{figure}[h]
    \centering
    \includegraphics[width=\textwidth]{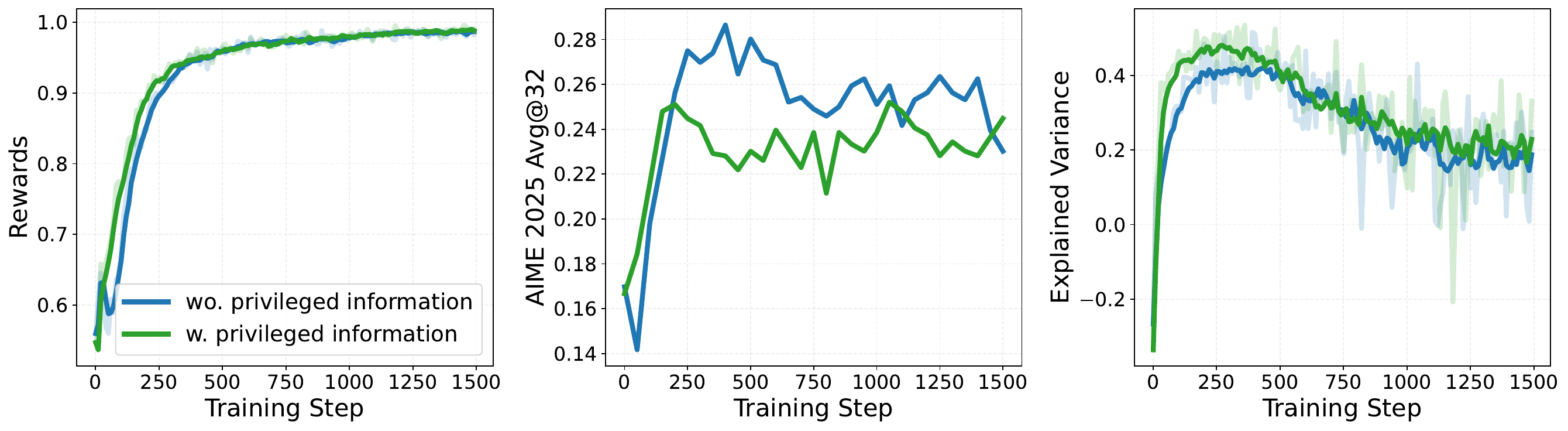}
    \caption{Effect of giving the reference answer only to the critic. Privileged input accelerates training-reward improvement (left) and increases explained variance (right). Its AIME 2025 avg@32 rises faster but peaks earlier (middle), revealing a greater risk of overfitting in this small-data setting.}
    \label{fig:privileged-information}
\end{figure}

The critic is needed only during training, so its inputs need not be identical to the policy's inputs. This observation parallels centralized training with decentralized execution in multi-agent RL~\citep{amato2024introduction}, including systems that expose hidden game state to a training-time critic~\citep{alphastar,wang2021scc}. We apply the same principle to reward-defining information in LLM RL.

Let $q(x)$ denote information used to evaluate responses to prompt $x$. For mathematical reasoning, $q(x)$ can be the reference answer. A privileged critic estimates
\begin{equation}
    V^\mu_\phi(s_t,q(x))\approx
    \mathbb{E}_{\mu}\!\left[R(x,y;q(x))\mid s_t,q(x)\right].
    \label{eq:privileged-value}
\end{equation}
Because $q(x)$ is fixed by $x$, exposing it explicitly does not change the optimal value associated with a prompt. It can nevertheless reduce the approximation burden on a finite model. The rollout policy still receives only $x$ and the generated prefix.

As shown in \Cref{fig:privileged-information}, using privileged information leads to faster and more stable training rewards, as well as higher explained variance. However, it also highlights a risk of overfitting: validation performance begins to decline earlier, despite improving more rapidly during the initial stage. 


\begin{figure}[h]
    \centering
    \includegraphics[width=\textwidth]{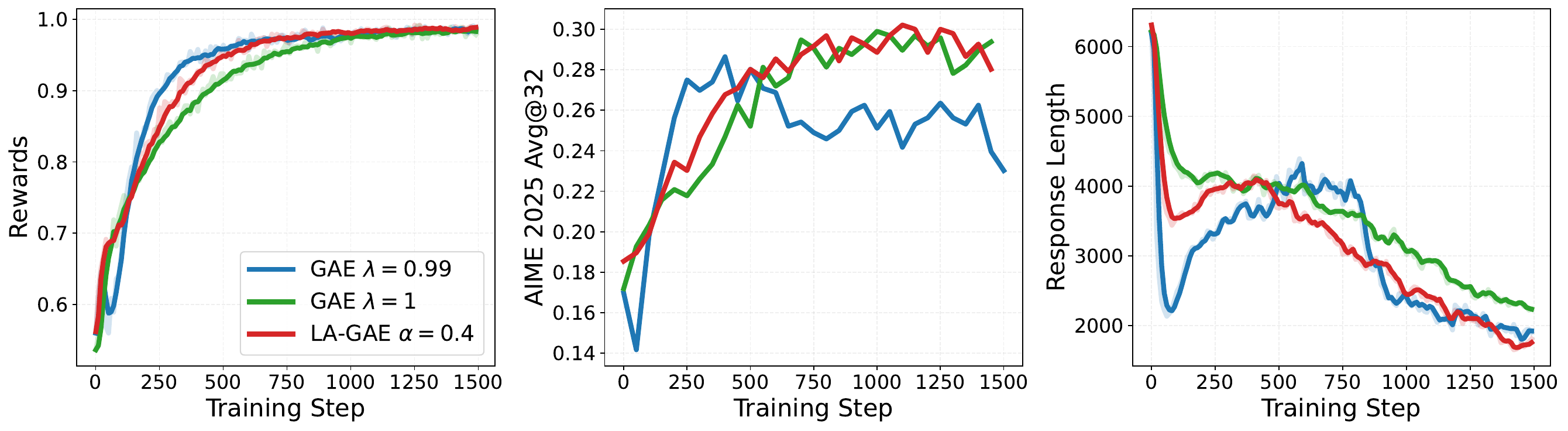}
    \caption{Effect of length-adaptive GAE. A fixed $\lambda_{\pi}=0.99$ attains rapid reward improvement (left), but its AIME 2025 avg@32 declines after the initial peak (middle). LA-GAE with $\alpha=0.4$ retains better training efficiency than $\lambda=1$ while mitigating the overfitting risk.}
    \label{fig:length-adaptive-gae}
\end{figure}

\subsection{Step 6: Adopting Length-Adaptive GAE}
\label{sec:lagae}

A fixed $\lambda_{\pi}<1$ gives the terminal reward exponentially less weight in longer responses. For a token at position $t$, the coefficient on the terminal TD residual is proportional to $\lambda_{\pi}^{T-t}$. When $T-t$ is large, early-token advantages depend primarily on bootstrapped critic residuals, and systematic critic error can dominate the policy signal.

Following VAPO and SAO~\citep{yue2025vapo,hou2026sao}, we use length-adaptive GAE\@. For a response of length $L=|y|$, the policy parameter is
\begin{equation}
    \lambda_{\pi}(L)=1-\frac{1}{\alpha L},
    \label{eq:length-adaptive-gae}
\end{equation}
where $\alpha>0$ controls the bias--variance trade-off. The coefficient of the terminal residual in the earliest-token advantage is approximately
\(
    \left(1-\frac{1}{\alpha L}\right)^{L}\approx\exp(-1/\alpha),
\)
which is nearly invariant to response length. We continue to use $\lambda_V=1$ for critic and do not use privileged inputs in this step.

As shown in \Cref{fig:length-adaptive-gae}, fixed $\lambda_{\pi}=0.99$ fits the training set fastest but exhibits a pronounced validation decline. Setting $\lambda_{\pi}=1$ avoids this decline at the cost of slower optimization. LA-GAE with $\alpha=0.4$ provides the best trade-off in this study. 

\section{Broader Evaluation}
\label{sec:evaluation}

We compare BPCO with group-based and critic-based baselines. The group baseline uses Dr.~GRPO~\citep{dr.grpo} with 16 responses per prompt. We reduce its number of distinct prompts so that all methods use the same total number of trajectories per iteration. The critic baseline includes two strong existing techniques: decoupled GAE with an unbiased Monte Carlo value target~\citep{yuan2025vcppo} and length-adaptive GAE~\citep{yue2025vapo}. It nevertheless retains an unbounded value head and batch-wise advantage normalization. BPCO differs from this critic baseline only by bounding value predictions and preserving raw advantages. All methods use DPPO for policy optimization~\citep{qi2026dppo}, which isolates advantage estimation from the choice of policy objective. \looseness=-1

\begin{figure}[h]
    \centering
    \includegraphics[width=\textwidth]{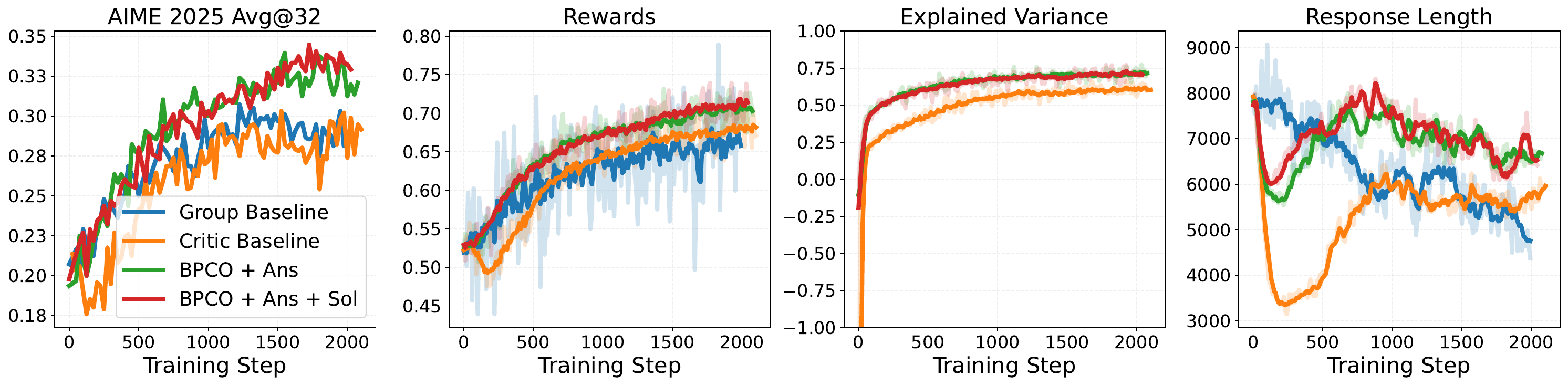}
    \caption{Results on a larger DeepScaleR dataset with DeepSeek-R1-Distill-Qwen-1.5B.}
    \label{fig:r1d-main-comparison}
\end{figure}

All critic-based methods sample one response for each prompt and warm up the critic during the first 15 iterations. We explicitly denote privileged variants by BPCO+Ans, BPCO+Sol, and BPCO+Ans+Sol. These variants provide the critic with the reference answer, the official solution, or both, respectively. The policy receives the same prompt as all baselines. For mathematical tasks, we report training reward, response length, and AIME 2025 avg@32. Explained variance is computed against the Monte Carlo target in \Cref{eq:unbiased-value-target}.

\subsection{Scaling to a Larger Dataset}
\label{sec:larger_dataset}

We first test whether BPCO scales beyond the small sanity-test dataset. We fine-tune DeepSeek-R1-Distill-Qwen-1.5B~\citep{deepseek_r1} on DeepScaleR~\citep{deepscaler2025}, which contains approximately 40.3K mathematical problem--answer pairs. Official solutions are available for about 7.3K problems. We allow generated responses of up to 24,000 tokens.

\Cref{fig:r1d-main-comparison} shows that our BPCO recipes consistently outperform both group-based and critic-based baselines, yielding clear improvements in training and validation performance. Compared with the critic-based baseline, BPCO achieves consistently higher explained variance throughout training, providing strong evidence that it learns a more accurate critic. These results demonstrate that BPCO critic remains effective on a substantially larger training set.

\begin{figure}[h]
    \centering
    \includegraphics[width=\textwidth]{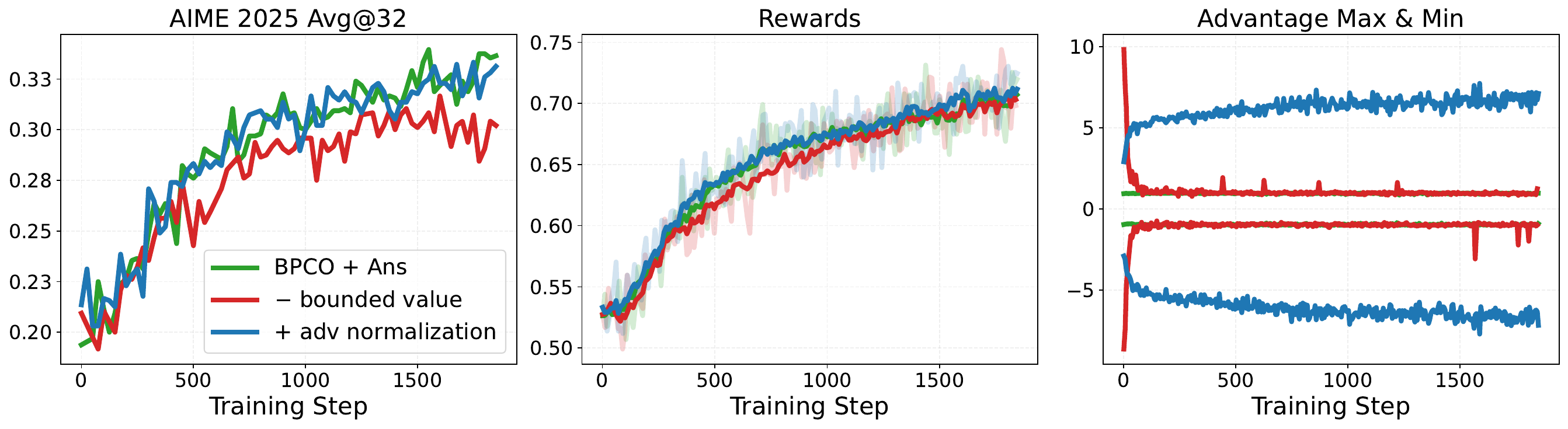}
    \caption{Ablation of bounded value prediction and batch-wise advantage normalization.}
    \label{fig:r1d-ablation}
\end{figure}

We next ablate our proposed techniques starting from a well-performing BPCO+\text{Ans} run. Specifically, we remove bounded value prediction and reintroduce batch-wise advantage normalization. The results are presented in \Cref{fig:r1d-ablation}.

\textbf{Bounded Value Prediction.}
Removing the value bound slows the improvement in training reward and reduces the AIME 2025 avg@32 score. Aligning the critic's output range with that of the return therefore remains beneficial, even when the training dataset is substantially larger.

\textbf{Removing Advantage Normalization.}
Reintroducing batch-wise advantage normalization causes the advantage magnitude to grow during training, although this effect is less pronounced than in the sanity test. Removing advantage normalization provides only modest performance gains in this setting, likely because training has not yet fully converged. Nevertheless, we recommend removing advantage normalization as a general-purpose default.

\begin{figure}[h]
    \centering
    \includegraphics[width=\textwidth]{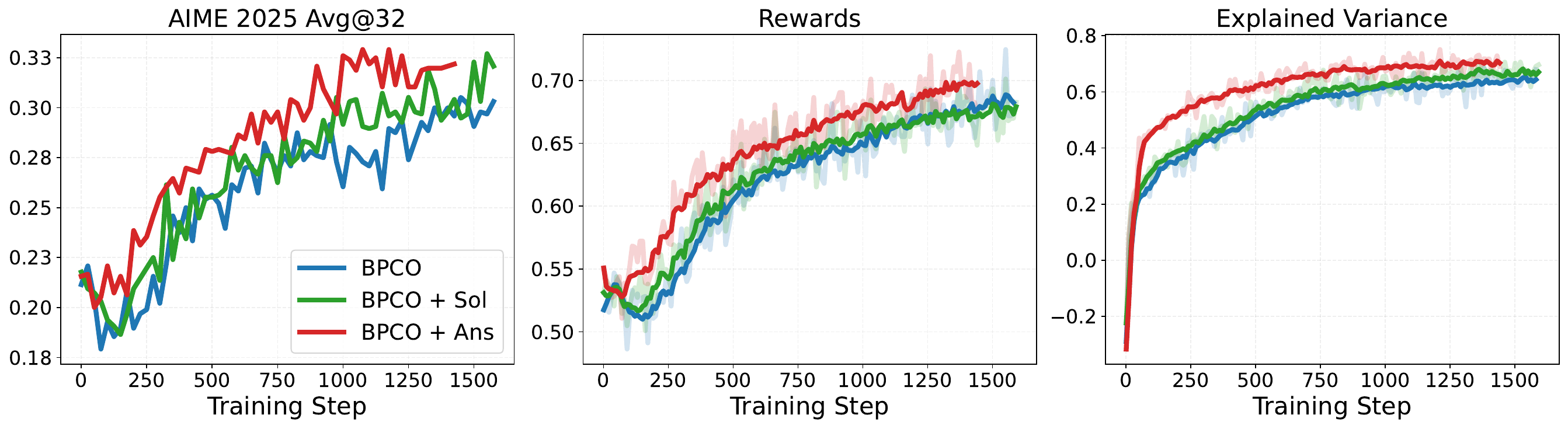}
    \caption{Ablation of privileged information for critic training.}
    \label{fig:r1d-privileged-information}
\end{figure}

\textbf{Privileged Information.}
\Cref{fig:r1d-privileged-information} ablates the effects of privileged information within our BPCO recipe. Providing the reference answer as privileged information leads to faster training, higher explained variance, and better AIME 2025 performance. Using the official solution also yields a modest improvement, even though only 7.3k of the 40.3k problems include this information. These results indicate that privileged information can substantially improve critic training when the dataset is sufficiently large and overfitting is not yet a concern.

\subsection{Scaling to Larger Models}
\label{sec:larger_model}

\begin{figure}[h]
    \centering
    \includegraphics[width=\textwidth]{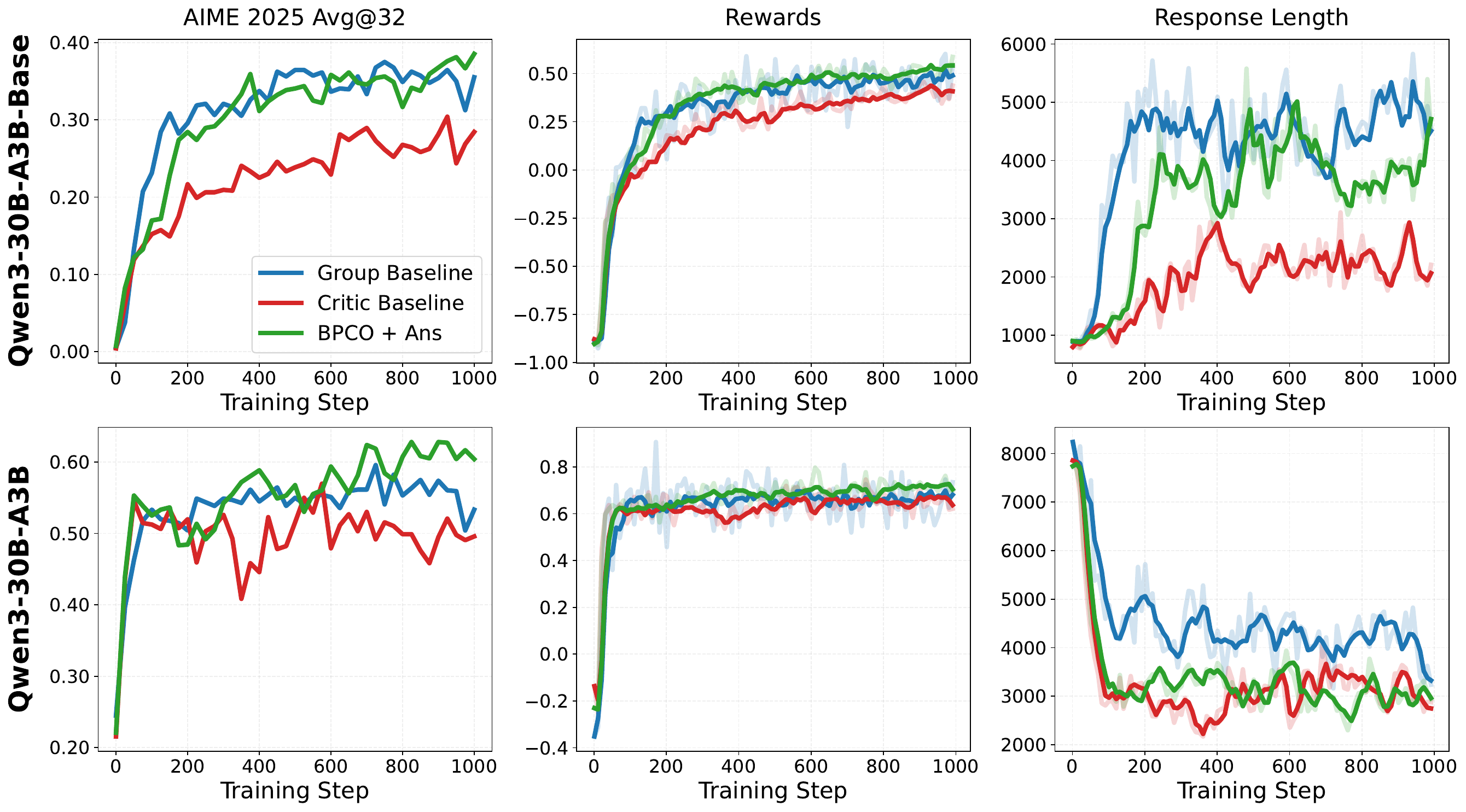}
    \caption{Results with larger MoE models on DAPO-Math-17k dataset.}
    \label{fig:moe-main-comparison}
\end{figure}

We further evaluate BPCO with Qwen3-30B-A3B-Base and Qwen3-30B-A3B~\citep{yang2025qwen3} on DAPO-Math-17K~\citep{yu2026dapo}. We set the maximum generation length to 12,000 tokens.

\Cref{fig:moe-main-comparison} shows that BPCO continues to improve upon the critic baseline at this scale. On Qwen3-30B-A3B, the critic baseline fails to improve AIME 2025 accuracy beyond the first 100 training steps, indicating unstable optimization. In both settings, BPCO achieves substantially higher AIME 2025 accuracy, demonstrating that it learns a more effective critic than the previous recipe.

Compared with the group baseline, BPCO performs better on Qwen3-30B-A3B and comparably on Qwen3-30B-A3B-Base. These results show that BPCO is a strong alternative to the widely adopted group-based method, without requiring group sampling~\citep{xu2026single,hou2026sao}.

\subsection{Rubric-Based Rewards}
\label{sec:rubrics}

We finally consider open-ended prompts evaluated by a rubric-based judge~\citep{gunjal2026rubrics}. We initialize both the policy and critic from Qwen3-4B-Base~\citep{yang2025qwen3} and train on OpenRubrics~\citep{liu2026openrubrics}. A frozen Qwen3-4B-Instruct-2507 judge scores each generated response against its prompt-specific reference rubric. The policy observes only the prompt. BPCO+Rubrics additionally exposes the same rubric to the training-time critic.

\begin{figure}[h]
    \centering
    \includegraphics[width=\textwidth]{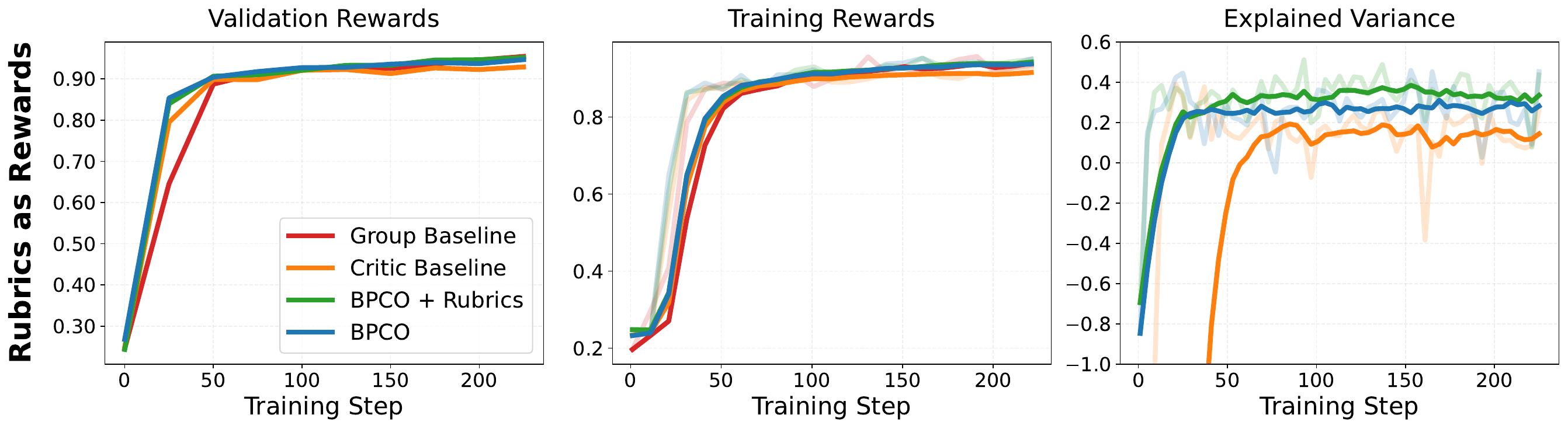}
    \caption{Results under rubric-based rewards.}
    \label{fig:rubrics-rewards}
\end{figure}

As shown in \Cref{fig:rubrics-rewards}, both BPCO variants learn faster than the group and critic baselines, although the group baseline eventually reaches comparable performance. The critic baseline achieves a slightly lower final reward, likely because of its suboptimal training recipe and low explained variance. Privileged information does not improve performance despite yielding higher explained variance, possibly because the task is relatively simple. Nevertheless, the stronger performance of BPCO without privileged information indicates that bounded value prediction and unnormalized advantages remain beneficial when rewards are provided by a rubric-based judge.

\section{Conclusion}

We studied why critic-based RL for LLMs can become unstable and assembled the resulting fixes into Best Practice Critic Optimization. BPCO combines DPPO with reward-range-bounded value predictions, unbiased Monte Carlo critic targets, unnormalized policy advantages, and length-adaptive GAE. It can additionally provide the training-only critic with privileged information, such as a reference answer, solution, or grading rubric, while leaving the policy's inputs unchanged. 

Across controlled sanity tests, larger datasets, 1.5B and 30B-A3B models, and rubric-based rewards, BPCO consistently improves a strong critic baseline. It also matches or exceeds group-based optimization while using a single response per prompt. Privileged information further improve critic learning when they provide useful reward context, but their gains are task dependent and can be offset by overfitting in small-data regimes. Overall, these results show that the critic itself is not an inherent weakness of LLM RL. When its output range, target, inputs, and induced policy signal are designed coherently, it provides a stable and efficient alternative to group-based estimation.

\paragraph{Limitations.}
Evidence is limited to mathematical and rubric rewards. BPCO assumes a known reward range, privileged variants require evaluator information, and critic training adds computation and memory not captured by trajectory-matched comparisons.






\clearpage
\bibliography{iclr2027_conference}
\bibliographystyle{iclr2027_conference}

\appendix

\end{document}